\def\year{2024}\relax
\documentclass[letterpaper]{article} 
\usepackage{aaai22}  
\usepackage{times}  
\usepackage{helvet}  
\usepackage{courier}  
\usepackage[hyphens]{url}  
\usepackage{graphicx} 
\def\UrlFont{\rm}  
\usepackage{natbib}  
\usepackage{caption} 
\DeclareCaptionStyle{ruled}{labelfont=normalfont,labelsep=colon,strut=off} 
\usepackage{algorithm}
\usepackage{algorithmic}
\UseRawInputEncoding

\usepackage{newfloat}
\usepackage{listings}
\floatstyle{ruled}
\newfloat{listing}{tb}{lst}{}
\floatname{listing}{Listing}
\usepackage{amsmath}
\usepackage{tabularray}

\renewcommand{\UrlFont}{\ttfamily\small}
\usepackage[final,inline,nomargin,index]{fixme} 

\FXRegisterAuthor{tl}{atl}{\color{blue}Taesung}
\FXRegisterAuthor{yp}{ayp}{\color{red}Youngja}
\FXRegisterAuthor{rg}{arg}{\color{cyan}Rishabh}
\FXRegisterAuthor{sh}{ash}{\color{purple}Shrayani}

\def\sec{Section}

\def\sec{Section}
\def\tab{Table}
\def\appdx{Appendix}
\def\fig{Figure}

\title{Towards Generating Informative Textual Description for Neurons in Language Models}
\author{
  Shrayani Mondal\textsuperscript{\rm 1}\equalcontrib\thanks{This work was done while the authors were at UMass Amherst.},
  Rishabh Garodia\textsuperscript{\rm 1}\equalcontrib\textsuperscript{\dag},
  Arbaaz Qureshi\textsuperscript{\rm 1}\equalcontrib\textsuperscript{\dag},
  \textbf{Taesung Lee\textsuperscript{\rm 2}\footnote{This work was done while the author was at IBM Research.}},
  \textbf{Youngja Park\textsuperscript{\rm 3}},
}
\affiliations{
    \textsuperscript{\rm 1}College of Information and Computer Sciences, UMass Amherst\\
    \textsuperscript{\rm 2}Bloomberg \\
    \textsuperscript{\rm 3}IBM Research\\
    \{smondal, rgarodia, squreshi\}@umass.edu, elca4u@gmail.com, young\_park@us.ibm.com

}

\usepackage{bibentry}

\begin{document}

\maketitle

\begin{abstract}
Recent developments in transformer-based language models have allowed them to capture a wide variety of world knowledge that can be adapted to downstream tasks with limited resources. However, what pieces of information are understood in these models is unclear, and neuron-level contributions in identifying them are largely unknown. Conventional approaches in neuron explainability either depend on a finite set of pre-defined descriptors or require manual annotations for training a secondary model that can then explain the neurons of the primary model. In this paper, we take BERT as an example and we try to remove these constraints and propose a novel and scalable framework that ties textual descriptions to neurons. We leverage the potential of generative language models to discover human-interpretable descriptors present in a dataset and use an unsupervised approach to explain neurons with these descriptors. Through various qualitative and quantitative analyses, we demonstrate the effectiveness of this framework in generating useful data-specific descriptors with little human involvement in identifying the neurons that encode these descriptors.
In particular, our experiment shows that the proposed approach achieves
75\% precision@2, and 50\% recall@2.
\end{abstract}

\section{Introduction}\label{sec:intro}



Recent breakthroughs in transformer-based language models have shown their capability of understanding and encoding diverse types of information,
and even performing emergent abilities as well as fine-tuning with a small amount of data for various tasks \cite{https://doi.org/10.48550/arxiv.1706.03762, https://doi.org/10.48550/arxiv.2005.14165,emergent}.
However, the low-level semantics of their most fundamental computational blocks still remain largely unknown.
This black-box nature of neural networks has led to a fast-growing and expansive area of research trying to understand how and what a model learns internally.




Providing a neuron level understanding through a \emph{descriptor}, such as ``general positive sentiment'', ``the subject is customer service'' or ``contains abusive language'', allows for understanding the inner working of a neural network on how it reaches a prediction, identifying biased or offensive neurons, applying artificial neurosurgery, and conforming to regulations~\cite{schick2021selfdiagnosis}. Increasing concerns on language models demands an explanation of model predictions~\cite{bowman2023things, ma2023oops}.
Many efforts are made to explain the model at a non-neural level.
For example, visualizing the relations and saliency of the input and output is one way to discover the input-output correlation as demonstrated by \citet{https://doi.org/10.48550/arxiv.1506.02078} and \citet{https://doi.org/10.48550/arxiv.1312.6034}. However, it does not show how various inputs are interpreted by neurons. Some other approaches prompt the model itself to add explanations alongside the outputs. A prominent example is a recent work by \citet{https://doi.org/10.48550/arxiv.2004.14546} that makes the T5 model \cite{https://doi.org/10.48550/arxiv.1910.10683} produce reasonings along with predictions for various downstream NLP tasks. This method can suffer from spurious explanations and could potentially make up a reasonable-sounding explanation instead of providing a truly accurate description of its causal decision-making process. In contrast, directly looking into what neurons get activated for a given input and what these neuron activations indicate can provide a deeper understanding of the internal logic.


Existing approaches tackled this problem of explaining neurons under a vision model setting to automatically label neurons with natural language descriptions.  \citet{DBLP:journals/corr/abs-2201-11114} and \citet{bau2017network} identify neurons that activate on specific image patches and proceed to label these neurons with descriptors.
However, the existing approaches are not easily applicable to the text domain for several reasons. First, due to a large variety of possible descriptors, it is not feasible to manually list an exhaustive set of candidate descriptors to identify among neurons. Second, reading and manually annotating text to tag them with a large set of descriptors can be more labor-intensive compared to identifying objects in an image. Third, the number of neurons in such language models can be very large, with the recent benchmarks reaching up to an order of billion parameters, making the process of manually annotating them resource intensive.

We propose to provide neuron-level explainability for language models using an unsupervised approach with minimal human interventions. In particular, we take advantage of open-source generative language models to generate meaningful descriptors and use classical NLP techniques such as clustering to obtain the candidate descriptors. We then automatically assign them to neurons of a widely used transformer-based model, which helps us to understand their behavior and contribution to various downstream tasks.

Our experiments take BERT as an example
and we evaluate our approach with BERT~\cite{https://doi.org/10.48550/arxiv.1810.04805} using the Amazon review dataset\footnote{https://s3.amazonaws.com/amazon-reviews-pds/readme.html}.
Our framework automatically extracts 23 candidate descriptors from the review dataset that highlight important aspects that a consumer might be interested in while buying a product.
Our evaluation involving human annotation and LLMs shows that
our approach achieves
75\% precision@2, and 50\% recall@2.
on the task of correctly tagging neurons with appropriate descriptors.
Also, the approach shows high consistency of 95\% Jaccard similarity when tested for two disjoint datasets from the same distribution, showing the descriptors are not spurious.

\section{Related Work}\label{sec:related}
The explainability of neural networks has been studied from many different angles, as described in ~\citet{https://doi.org/10.48550/arxiv.2108.13138}.
Some approaches are input-based, connecting the input and the model prediction directly, often considering the model as a black box.~\citet{smilkov2017smoothgrad} proposed an input-based approach by computing the correlation of input features to predictions and then visualizing them.
\citet{dalvi2018grain} and \citet{torroba-hennigen-etal-2020-intrinsic} developed input-based approaches that provide human-readable descriptions which can further explain how the raw input features are understood by the model.
This can be useful, especially when the input features are not immediately interpretable.
\citet{https://doi.org/10.48550/arxiv.2205.07237} leverages human-in-the-loop to assign such a descriptor. These input-based approaches often describe the model as a black box component and do not focus on its internal workings. 

Another direction of approaching this problem focuses on explaining neurons instead, and they try to understand a neural network based on neuron activations.
\citet{https://doi.org/10.48550/arxiv.1602.08952} qualitatively analyze various linguistic properties encoded in the hidden dimensions (neurons) of RNN models. By extracting 5-gram concepts for every neuron and analyzing them, they were able to identify interesting phenomena like neurons predictive of a grammatical function or individual neurons that become highly sensitive towards contexts with syntactic patterns.

In another attempt to understand the representations of CNNs trained on language tasks, \citet{https://doi.org/10.48550/arxiv.1902.07249} selected $k$ most, activating sentences for every neuron and discovered the underlying concepts by parsing them using parse trees. They generated synthetic sentences that highlighted a concept, often with the risk of these synthetic sentences being ungrammatical and/or with repetition. 

\citet{mu2021compositional} show how neurons are not just simple feature detectors, but rather operationalize complex decision rules composed of multiple concepts. The authors build off of the works by \citet{bau2017network} but instead show how one neuron's behavior should be explained as the composition of concepts. They start out by pre-defining a concept inventory and producing explanations that best explain the neurons' behavior over these concepts. This is done by considering Intersection over Union score as a measure. The authors go a step further and create combinatorial compositions of explanations ($\mathcal{C}_{1}$ \textbf{OR} $\mathcal{C}_{2}$ \textbf{AND} $\mathcal{C}_{3}$).

\citet{DBLP:journals/corr/abs-2201-11114} 
proposed a global, neuron-based approach to computer vision models.
They demonstrate that some neurons in computer vision models are highly capable of identifying semantic and structural features of the input by linking neurons with textual descriptors. They start out by representing each neuron with what the paper calls ``exemplar sets." These are collections of input image patches that maximally activate a neuron. They generate natural language descriptions of individual neurons describing the common characteristics of these image patches by optimizing \emph{Pointwise Mutual Information (PMI)} of the candidate descriptor text and the exemplar sets. To model PMI, they use a variation of an image captioning model, and a two-layer LSTM language model, both trained on their manually annotated MILANNOTATIONS dataset.

\begin{figure*}[!ht]
    \centering
    \includegraphics[width=\textwidth]{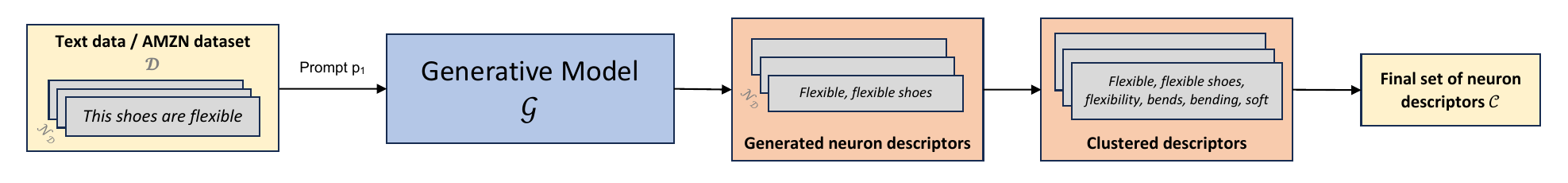}
    \caption{The procedure to generate candidate descriptors. The descriptors can be generated for a large dataset using generative LLMs. They are clustered to reduce different expressions referring to the same meaning.}
    \label{fig:concept_gen_process}
    \end{figure*}
    
    \begin{figure*}[ht]
    \centering
    \includegraphics[width=\textwidth]{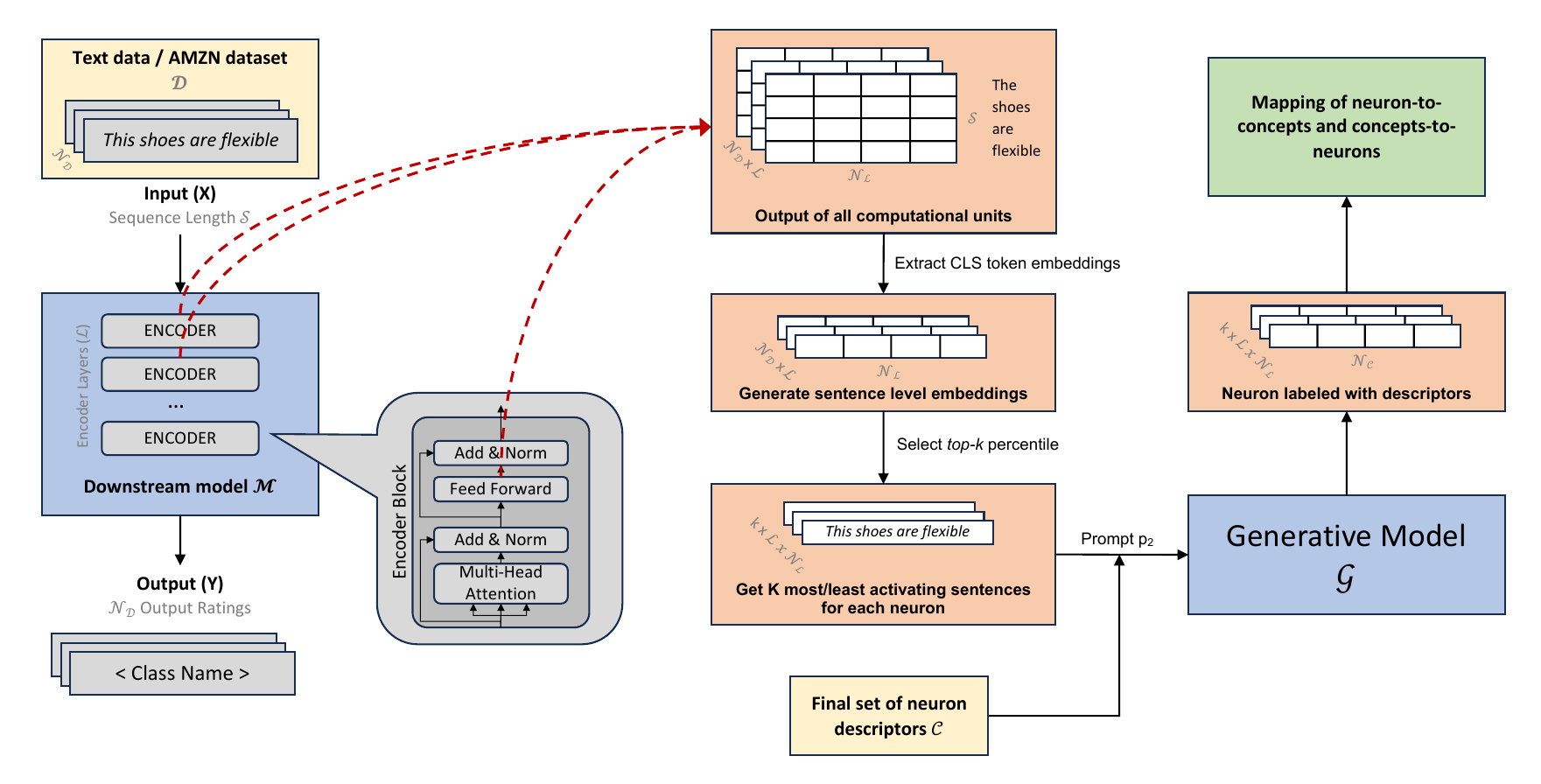}
    \caption{Proposed process flow for generating descriptors for neurons in LLMs being used for any downstream task.}
    \label{fig:process_fig}
\end{figure*}

These existing approaches either
focus on correlation, lacking natural language descriptors,
or require heavy human intervention of manually providing descriptions of neurons or input sentences.
Our framework attempts to overcome these shortcomings to make the framework automatic, unsupervised, and widely applicable.

Our approach is inspired by \cite{DBLP:journals/corr/abs-2201-11114} and \cite{mu2021compositional} where we attempt to overcome the shortcomings by substituting supervised components with unsupervised approaches, replacing data-specific techniques with more general and global ones and provide a list of descriptors for each neuron of the form ($\mathcal{C}_{1}$ \textbf{AND} $\mathcal{C}_{2}$ \textbf{AND} $\mathcal{C}_{3}$) instead of restricting one neuron to a single descriptor.



\section{Approach}\label{sec:approach}

Our framework tries to describe the neurons in a text-based deep learning model $\mathcal{M}$ with natural language descriptors. We take an approach using a set $\mathcal{D}$ of sentences with their descriptors, feeding the model $\mathcal{M}$ with $\mathcal{D}$, and analyzing the neuron activations to apply the descriptors from the sentences to the neurons (\sec~\ref{sec:method:neurons}, 
summarized in \fig~\ref{fig:process_fig}).

However, this process requires a dataset with descriptors which has been manually created in the existing works. We leverage one or more generative language models to find candidate descriptors that can be used for the model $\mathcal{M}$ using the dataset $\mathcal{D}$ (\sec~\ref{sec:method:candidates}, summarized in \fig~\ref{fig:concept_gen_process}), and assign such descriptors to sentences in $\mathcal{D}$ (\sec~\ref{BinaryClassification}).

\subsection{Identifying Candidate Descriptors} \label{sec:method:candidates}

Different models encode different pieces of information, which also depend on the input data. Thus, the descriptors depend on $\mathcal{M}$ and $\mathcal{D}$ and these contraints create a scope for candidate descriptors. 
That is, if the model does not have any neuron to recognize a certain pattern in the input, a descriptor for the pattern is not useful. Also semantically similar descriptors indicating the same pattern might dilute the reaction of a neuron to multiple descriptors.
Here, we explain how we create a list of candidate descriptors for $\mathcal{M}$ that can be obtained through $\mathcal{D}$ (\fig~\ref{fig:concept_gen_process}).



\begin{figure}[h!]
    \centering
    \frame{\includegraphics[scale=0.34]{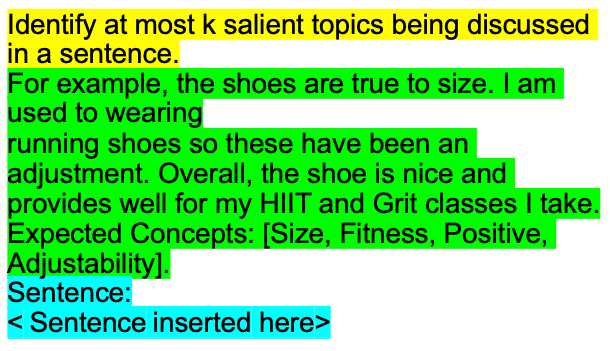}}
    \caption{A prompt template for identifying candidate descriptors. It is made up of the task (yellow), 1-shot example (green) and an input sentence in question (blue).}
    \label{fig:p1_template}
\end{figure}

Suppose we have a dataset $\mathcal{D}$ of sentences $d_{1}, \ldots, d_{|\mathcal{D}|}$.
For each $d_{i} \in \mathcal{D}$, we ask an instruction fine-tuned LLM $\mathcal{G}$, such as FLAN-T5 XXL, with a prompt $p_{1}$ in context of $d_{i}$ to obtain the descriptors of $d_{i}$. To further aid the generation task and direct generation towards the desired format, a 1-shot example $e$ is provided along with $p_{1}$. 
This 1-shot example is randomly chosen from $\mathcal{D}$ and the same example is used for all $d_{i} \in \mathcal{D}$.
\fig~\ref{fig:p1_template} shows an example of $p_{1}$.
To collect a diverse set of descriptors, we can use multiple such LLMs and use the union of the responses.


Due to the very nature of generative tasks, different descriptors indicating the same meaning can be generated across $\mathcal{D}$. Thus, we cluster them to identify those with similar meanings. In our implementation, we apply Fast Community Detection algorithm\footnote{\url{https://www.sbert.net/docs/package_reference/util.html}} using Sentence-BERT~\cite{reimers-2019-sentence-bert} to find descriptor clusters. We define a set $C$ of clusters containing semantically similar descriptors and use them to describe neurons later. We can choose a representative descriptor for each of the clusters for readability. This can be done manually or using automated cluster representation approaches such as \cite{poostchi-piccardi-2018-cluster}.
For our experiments, we assign cluster labels manually, as shown in the following examples.
\begin{itemize}
    \item “pigmentation: 5/5”, “pink color”, “hue”, “colored”, “blue" can be represented by \textbf{Color}.
    \item “sounds and looks great”, “audible”, “loud noise”, “sound absorption” can be represented by \textbf{Sound}.
\end{itemize}
More examples are presented in \tab~\ref{tab:examples_clustering}.

\subsection{Obtaining Descriptors for Sentences}\label{BinaryClassification} 
\begin{figure}[!ht]
    \centering
    \frame{\includegraphics[scale=0.34]{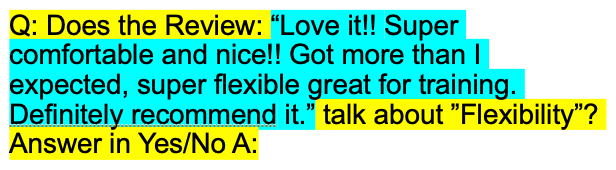}}
    \caption{A prompt template for obtaining descriptors for sentences. It is made up of the task (yellow), and an input sentence $d_i$(blue).}
    \label{fig:p2_template}
\end{figure}

Based on this concise set $C$ of candidate descriptors, we have to label the corpus $\mathcal{D}$ that we feed to the model later.
This is a multi-class text classification problem, and we exploit generative LLMs. For every sentence $d_{i} \in \mathcal{D}$ and every descriptor $c \in C$, we ask $\mathcal{G}$ with a prompt $p_{2}$ to answer if the descriptor $c$ is applicable to $d_{i}$. We format $p_{2}$ so that the generated text is limited to the words ``Yes'' or ``No''. An example can be seen in \fig~\ref{fig:p2_template}. This generates a $|\mathcal{D}| \times |C|$ binary matrix $\mathcal{B}$ (1 for ``Yes'' and 0 for ``No'') representing the descriptors applicable to the sentences.

\subsection{Explaining Neurons with Descriptors} \label{sec:method:neurons}

In this step, we analyze how the target model $\mathcal{M}$ and its neurons react to the input sentences to find the descriptors for the neurons, as shown in \fig~\ref{fig:process_fig}.
We feed each sentence from our dataset $\mathcal{D}$ to $\mathcal{M}$ and keep track of the top sentences that activate a neuron. As the output of one neuron for one sentence is a vector, we extract one dimension of this vector that allows us to rank the sentences.
These top sentences are called an \emph{exemplar set}, that contains sentences highly activating the neuron.
We assign the common descriptors found in the exemplar set to the neuron, indicating that the neuron has learned to detect this common pattern.

More specifically, we forward pass all sentences in $\mathcal{D}$ through $\mathcal{M}$ and record the sentence level activations for each neuron-sentence pair. Then, for each neuron $n_i$, the top $k$-percent sentences with the highest activations are taken as an exemplar set $\mathcal{E}_{i}$ for $n_{i}$.
Then, we assign a descriptor $c$ to $n_{i}$ if it appears frequently among $\mathcal{E}_{i}$.
That is, we compute the descriptors $C_{n_i}$ of neuron $n_{i}$ as follows.
\begin{align*}
    C_{n_i} = \{c \in C_{\mathcal{E}_{i}}\; |\; f(c) > t \}
\end{align*}
where $C_{\mathcal{E}_{i}}$ is the list of all occurrences of the descriptors $c$ in the exemplar set $\mathcal{E}_{i}$; $f(c)$ is the $\text{percentage frequency of $c$ in $C_{\mathcal{E}_{i}}$}$ and $t$ is a composition threshold.
An inverse mapping can also be created to get a  list of neurons tagged to a certain descriptor.

\section{Experimental Setup}\label{sec:experiments}
In this section, we describe our experiments including parameters, models, and datasets used as well as some intermediate results.
We also compare different options available at different stages of our method.
We show our quantitative and qualitative evaluation results in \sec~\ref{sec:results}.

\begin{figure*}[!ht]
    \centering
    \includegraphics[width=\textwidth]{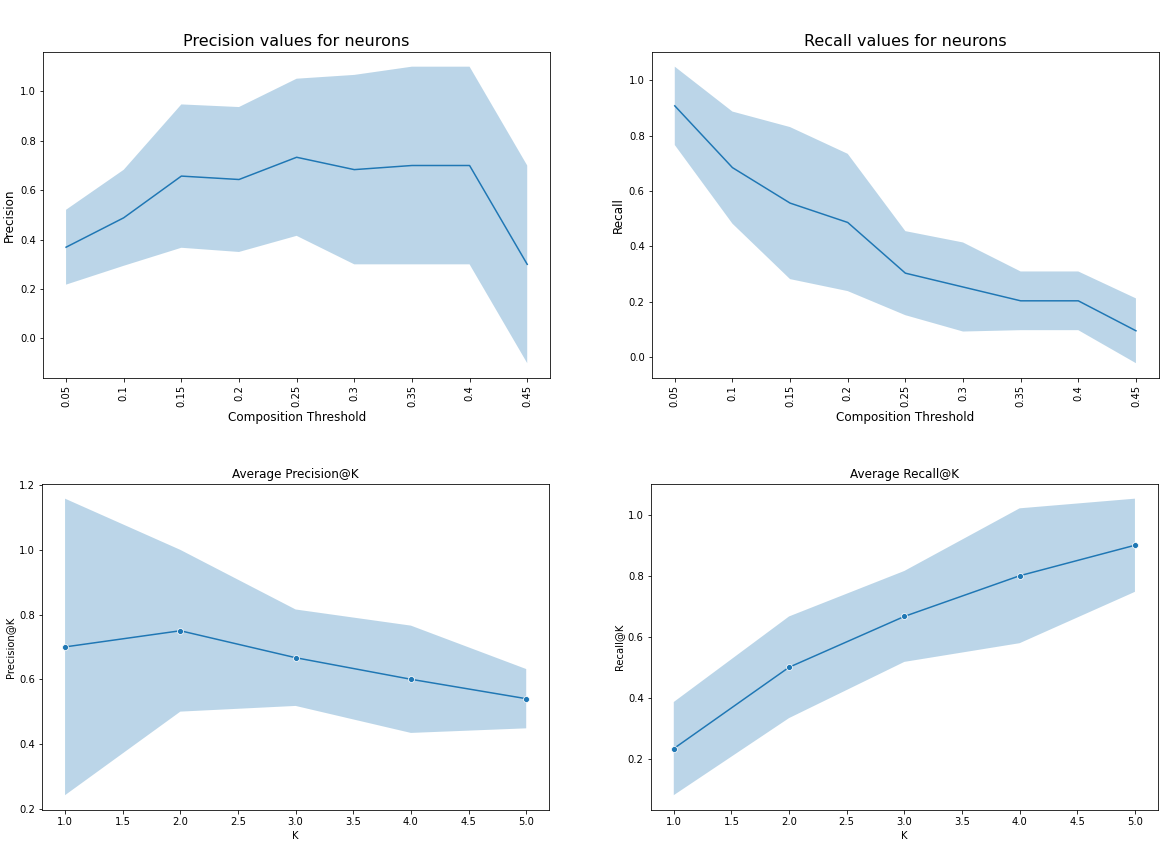}
    \caption{Precision and Recall. The shade shows standard deviation. Top Left: Precision vs Composition Threshold, Top Right:  Recall vs Composition Threshold, Bottom Left: Average Precision@K vs K, Bottom Right: Average Recall@K vs K.}
    \label{fig:All_prec_recall}
\end{figure*}

\subsection{Models and Parameters}


For our experiments, 
we study the BERT-base-uncased model without fine-tuning as model $\mathcal{M}$ to explain\footnote{https://huggingface.co/bert-base-uncased}. It consists of $\mathcal{L}=12$ encoder blocks each containing a multi-head attention layer, followed by ADD \& NORM operation layer, followed by fully-connected layer followed by another ADD \& NORM operation layer~\cite{https://doi.org/10.48550/arxiv.1810.04805,https://doi.org/10.48550/arxiv.1706.03762}. We work with the activation values extracted from the fully-connected layer before the ADD \& NORM operation is applied. The output of a fully-connected layer in BERT-base-uncased model is of the shape $\mathcal{N}_{\mathcal{L}} \times S$, where $\mathcal{N}_{\mathcal{L}}$ is the number of neurons in that layer and $S$ is the sequence length. For every sentence, only the \textit{[CLS]} token activations are extracted, therefore, $S=1$. This implies that one forward pass on one sentence generates $\mathcal{L} \times \mathcal{N}_{\mathcal{L}}$, that is, $12 \times 768 = 9216$ total activation values. 


For the instruction fine-tuned large language model $\mathcal{G}$, we experimented with 2 generative models --- FlanT5 XXL \cite{https://doi.org/10.48550/arxiv.2210.11416} and the 4th iteration English supervised-fine-tuning (SFT) model of the Open-Assistant project based on a Pythia 12B \cite{OpenAssistant/oasst-sft-4-pythia-12b-epoch-3.5} (henceforth referred to as Pythia model). We choose these models because they are available in open-source and have good-quality of output, so we can apply on a large amount of sentences in $\mathcal{D}$. We use a union of descriptors from both of these models to obtain a diverse set of descriptors (refer \tab~\ref{tab:concept-generation} for examples). We find, on one hand, the descriptors from FlanT5 XXL are more accurate than the Pythia model, but on the other hand, the Pythia model generated descriptors of a wider variety and contained some fine-grained descriptors that were missed by the FlanT5 XXL model.


Finally, we have parameters $k$ and $t$. To get the exemplar sets $\mathcal{E}_{i}$ for each neuron $n_i$, we sort the sentences in decreasing order of their activation values and select the top $k=1$ percent from them as described in \sec~\ref{sec:method:neurons}. We limit k=1 throughout the paper giving us an exemplar set of 435 reviews. With 10 to 15 token per review, it is assumed that we will be well within the permissible input token window of $\mathcal{G}$. We vary the composition threshold $t$ in order to evaluate its impact in \sec~\ref{sec:exp:jaccard}.

\subsection{Dataset}


Our experiment leverages the Amazon Product reviews dataset~\cite{5-core-paper}
to focus on a set of descriptors related to products sold on the e-commerce website.
That is, in \sec~\ref{sec:approach}, we explained that in order to label neurons with the descriptor that they activate on, we need to have a dataset of sentences and their corresponding sentence descriptors.
We use this dataset due to its more focused nature, and apparent and comprehensible features discussed in the corpus,
which makes the human annotation task easier.

This dataset consists of Amazon reviews from 18 product categories like electronics, furniture and cosmetics. It has more than 130 million customer reviews spanning over a period of 2 decades (1995 - 2015).
We randomly select a maximum of 5,000 reviews from each of the 18 product category from the 5-core subset\footnote{https://nijianmo.github.io/amazon/index.html}
of this dataset to create a corpus of 111,611 reviews.
A 2-step filtering is applied to this to remove reviews that have less than 10 or more than 200 words and retain reviews that are in English only.
This brings us to a final corpus of 86,948 reviews.
From now on, we refer to this subset as ``AMZN'' dataset.

We split this dataset into two subsets of equal sizes --- calibration set and validation set. We do this to be more confident of the neuron descriptors that we obtain. That is, if we obtain the same neuron descriptor consistently from both the calibration set and the validation set, it is likely that the neuron encodes the descriptor described by that descriptor. 1 percent of each of these two sets gives us an exemplar set of size 435 reviews, and it is large enough to select a descriptor based on its frequency.
This provides a 50-50 split of reviews and maintains equal distribution for each descriptor which can be seen in  \fig~\ref{fig:DataDistribution}.

\subsection{Sentence Annotation with Descriptors}

\begin{table*}[t]
\centering
\caption{Examples of descriptor generation using OpenAssistant SFT Pythia model and FlanT5 XXL model}
\small
\label{tab:rawdescriptors}
\begin{tblr}{
  width = \linewidth,
  colspec = {Q[180]Q[120]},
  vline{-} = {-}{0.08em},
  hline{-} = {-}{0.05em},
}
\textbf{Reviews} & \textbf{Descriptors generated from OpenAssistant and FlanT5 XXL model}\\
Relieved my Plantar Fascitis for 3 Days. Then the unbearable pain returned in full force. These were recommended by my Podiatrist. & 	'bad', 'pain','relief', 'plantar fasciitis', 'unbearable pain', 'podiatrist', 'relief', 'recommended by podiatrist' \\
I purchasaed a new dryer and did not want to reuse the cord from my old unit. This unit installed in a pretty straight forward manor. Quality was as expected. No Complaints & 'easy to install', 'easy to use', 'quality', 'no complaints', 'reusable cord' \\
I purchased the Kindle edition which is incredibly handy, particularly when traveling.  Melissa Leapman is always dependable for providing those wonderful necessities for knitters. &  'reliable', 'necessary', 'knitters', 'dependable', 'knitting necessities', 'handy' \\
This game is a bit hard to get the hang of, but when you do it's great. & 	'game', 'fun', 'hard', 'enjoyment', 'learning curve', 'gameplay', 'challenging', 'frustrating', 'persistent', 'difficulty' \\
Awesome heater for the electrical requirements! Makes an awesome preheater for my talnkless system & 'user experience', 'positive', 'awesome', 'electrical requirements' \\
Keeps the mist of your wood trim and on you. Bendable too. & 'good', 'keeps the mist of your wood trim and on you', 'bendable' \\
\end{tblr}
\end{table*}

\begin{table*}[t]
\centering
\caption{Examples of clustered descriptors.}
\label{tab:clustering}
\small
\begin{tblr}{
  width = \linewidth,
  colspec = {Q[240]Q[80]},
  vline{-} = {-}{0.08em},
  hline{-} = {-}{0.05em},
}
\textbf{Descriptor Cluster} & \textbf{Representative Descriptor}\\
'simple and easy to use', 'such as ease of use', 'simple and straightforward', 'easy to pick up', 'basic instructions included', 'friendly user interface', 'versatile and easy to use', 'no easy to use', 'easy to use and learn', 'simple but nice', 'fairly easy' & Easy to use\\

'well-packaged', 'quality/service/packaging', 'bulk pack', 'second box', 'case', 'nice little case''the packaging was good too', 'quality of packaging', 'inaccurate packaging', 'delivery and packaging', 'wooden box', 'cardboard box' & Packaging \\

'determining size', 'attractive fit', 'nice fit', 'fit as expected', 'little', 'good width', 'not big enough', 'slim size', 'classic fit', 'fit and finish' & Size/Fit \\

'taste preferences', 'tasteful', 'flavor is just okay', 'sugar water', 'sweet but not overly so', 'tastes like cheese', 'sweet and spicy', 'delicious food', 'smooth and flavorful', 'rich flavor' & Taste/Flavour \\

'unhealthy skin', 'skin hygiene', 'great for sensitive skin', 'normal skin', 'skin imperfections', 'moisturizer', 'intense moisturizing', 'effective for pimples', 'acne free', 'makeup sponge', 'good shave', 'effective pre-shave', 'dry skin relief' & Skincare\\

'flavored tea', 'Tea', 'premium tea', 'thai iced tea', 'strong tea', 'Turkish coffee', 'coffee taste', 'great coffee', 'white tea', 'tea as a daily treat', 'diverse coffee flavors', 'foreign to coffee notes' & Beverage \\

\end{tblr}
\end{table*}

We process the AMZN dataset through the steps described in \sec~\ref{sec:method:candidates} and \sec~\ref{BinaryClassification}.
We get an initial set of descriptors for each of the review sentences using two prompt-based LLMs: Flan-T5 XXL and Pythia model that was fine-tuned on human demonstrations of assistant conversations collected through human feedback (\tab~\ref{tab:rawdescriptors}). We perform the clustering as described in \sec~\ref{sec:method:candidates} to obtain descriptor clusters and their representative descriptors such as shown in \tab~\ref{tab:clustering}. In our case, we obtain a list of 26 representative descriptors
and discard 3 of them from this list (["Positive", "Product Quality", "User Experience"]) (refer to \tab~\ref{tab:all_descriptors} for all representative descriptors and \appdx~\ref{appdx:clusters} for more examples). This is done as these 3 descriptors have broad connotation and have high activations for more than 80\% of neurons, skewing the evaluation metric, and making one of these descriptors an easy win for the proposed method. We are interested in analysing more specific and diverse descriptors. Based on the end-user's need, the final list of descriptors to work with can be updated as seen fit.

\section{Results and Analysis}\label{sec:results}

\subsection{Neuron Descriptor Evaluation}

Firstly we present the evaluation result of neuron descriptors $\{C_{n_i}\}$.
For quantitative analysis, we strategically sample $\mathcal{D}$ in the following way to get the labeled data:
First, for each of 23 candidate descriptors, we randomly choose 15 sentences that are tagged with the descriptor by Flan-T5 XXL, and another 15 that are not, leading to a final sample of 690 sentences. A human assessor manually annotates them again with Yes/No for each descriptor. We get 76\% precision and 94\% recall when the Flan-T5 XXL output is assessed against manual annotations for these 690 sentences.

To scale up the evaluation, we propose using ChatGPT as a proxy for human assessment. We compared the manual and ChatGPT annotation for those 690 reviews and achieved an inter-annotator agreement of \textbf{0.865} using Cohen's kappa score showing that ChatGPT annotations are reliable. We randomly selected 10 neurons, and tag their top 1\% activating sentences using ChatGPT with the list of 23 descriptors.
A total corpus of 4,350 sentences were labeled and then used as ground truth to analyze these 10 neurons by computing precision and recall on the task of tagging descriptors to neurons.




As shown in \fig~\ref{fig:All_prec_recall}, overall, we see similar results but with a lower recall likely due to biases in the LLMs. The varying evaluation parameters show expected effects. The top-left figure shows a general increase in the precision of the predicted descriptors as we increase the composition threshold.
A higher composition threshold value would encourage the framework to pick only prominent descriptors for the top activating sentences and discard anything that falls below the threshold.



In order to eliminate the dependency on the composition threshold, we use top-K descriptors. We first limit the tagged descriptors in the ground truth to top-3 per neuron instead of using a composition threshold. For evaluating the output of tagging from our framework, we use top-K descriptors per neuron to compute Precision@K and Recall@K for different values of K.
When averaged over 10 neurons, Precision@K peaks when we consider the top-2 descriptors (see \fig~\ref{fig:All_prec_recall}, bottom-left) resulting in an average precision@2 of 75\%, and the average Recall@2 value is 50\%. (see \fig~\ref{fig:All_prec_recall}, bottom-right).


\subsection{Relations Among Descriptors}

Once descriptors are obtained, we can compute the Pearson Correlation between the descriptors using the $|\mathcal{D}| \times |\mathcal{C}|$ binary matrix $\mathcal{B}$ from \sec~\ref{BinaryClassification}.
We can find strong dissimilarity between
``Negative'' and ``Positive'',
and high similarity between (``Taste/Flavour'' and ``Beverage''), (``Texture'' and ``Fabric''), (``Graphics'' and ``Design''), and (``Gift/Present'' and ``Age Appropriate/For Kids''),
showing how they often co-occur in the related inputs.
More result can be found in \appdx~\ref{appdx:corr}.

\subsection{Neuron Descriptor Consistency} \label{sec:exp:jaccard}

\begin{figure}[t]
\centering
\includegraphics[scale=0.5]{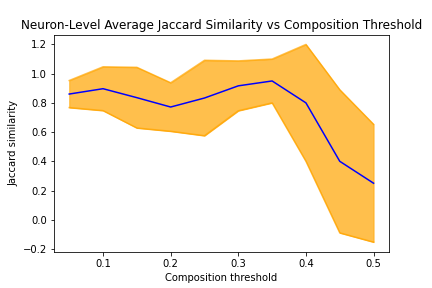}

\caption{Neuron-level Average Jaccard Similarity starts to drop as we increase the composition threshold. The shade shows standard deviation, which increases as the composition threshold increases.}
\label{fig:selected_neuron_avg_JC_vs_CT}
\end{figure}

\begin{figure}[t]
\centering
\includegraphics[scale=0.5]{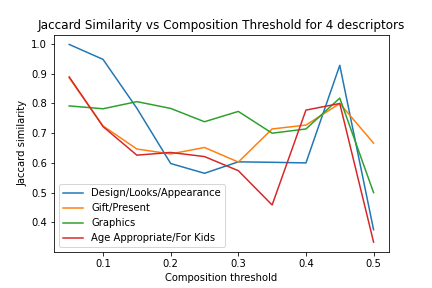}

\caption{Descriptor-level Jaccard similarity follows the trends from Neuron-level Average Jaccard similarity and drops as composition threshold increases.}
\label{fig:ConceptLevelJS}
\end{figure}

We also evaluate the consistency of the descriptor as an additional automated evaluation to check if the obtained neuron descriptors are spurious.
For a specific neuron $n_{i}$, we get the 2 sets of descriptors $C_{n_i}$ and $C^{'}_{n_i}$ assigned to it from the calibration and validation sets, respectively. We analyze the consistency of these 2 sets by computing the Jaccard similarity between them using this formula:
\begin{align*}
    J(C_{n_i}, C^{'}_{n_i}) = \frac{|C_{n_i} \cap C^{'}_{n_i}|}{|C_{n_i} \cup C^{'}_{n_i}|}
\end{align*}


As it can be seen in  \fig~\ref{fig:selected_neuron_avg_JC_vs_CT}, the average Jaccard similarity for the 10 selected neurons peaks at a composition threshold of 0.35. We obtain a Jaccard similarity of 0.95 at this value of the composition threshold. When we increase the composition threshold from 0.05 to 0.35, we are eliminating low-quality descriptors, as their percentage frequency is low. But when we increase the composition threshold beyond 0.35, we see a drop in the size of the descriptor sets obtained using both the calibration and the validation set. Hence, the Jaccard similarity is smaller for these composition threshold values.

Similarly, when we invert this mapping of neuron to descriptors, we obtain a distinct set of neurons tagged to each descriptor. \fig~\ref{fig:ConceptLevelJS} shows the Jaccard similarity in neuron sets for four descriptors when we vary the composition threshold. Intuitively, a low composition threshold should lead to low-quality neuron descriptors and many neurons would get tagged to many descriptors. This leads to high Jaccard similarity but low-quality descriptors.
We notice that this descriptor-level Jaccard similarity peaks at a composition threshold of 0.45, indicating that we get a maximum overlap between the calibration and validation set neurons. We then see a sharp drop when we increase the composition threshold to 0.5.
This plot also shows that at a composition threshold of 0.45, a neuron is often tagged with similar sets of descriptors, using both the calibration set and the validation set. 

\section{Limitations}\label{sec:discussion}
Our experiments focus on discriminative models for $\mathcal{M}$. While we believe a similar framework can be used for a generative model, our evaluation is limited to a discriminative model. We think explaining neurons of a generative LLM would provide us even more interesting result.
In this paper, we exclusively focused on sentence level semantics by considering the activation of the \textit{[CLS]} tokens which is often used to represent the overall semantic of the input sentence. Therefore, our approach focuses on the presence of information in the sentence regardless of the token position, captured by individual neurons. On the other hand, applying a similar approach while targeting token level semantics may provide additional insights on more neurons as well as generative models where the first token does not pay attention to other tokens appearing after it.

The effectiveness and accuracy of our approach are heavily dependent on the ability of generative LLMs to discover the inherent concepts in the sentences. Despite the high Cohen's kappa in our evaluation, when dealing with a textual corpus of large input text like stories or legal documents, these LLMs may not provide an exhaustive list of descriptors. Also, they are prone to hallucinations which can bring in noise as well. 

The textual descriptors that we use in the experiments section are limited to words or short phrases instead of lengthier sentences. We see potential in this approach and believe that our work can be extended by adapting the prompts to obtain sentence-like descriptors for neurons as well.

Getting ground truth descriptors for neurons is a labor-intensive task. We believe tagging more neurons manually or using ChatGPT could have benefited the evaluation process. Similarly, evaluation can be performed on different downstream models $\mathcal{M}$ and different datasets $\mathcal{D}$.

Other extensions of this work would include
expanding the number of neurons to work with, exploring other datasets or a mix of datasets, analyzing the effect of model fine-tuning on the generated descriptors.

\section{Conclusion}\label{sec:conclusion}
Through this paper, we provide a novel unsupervised framework to explaining neurons with human-interpretable descriptors in LLMs. We eliminate the requirement of starting out with an initial inventory of descriptors for neuron-level analysis and minimize human interventions. Our experimental results validate the potential of our approach with reasonable precision and recall. Our framework is scalable and can adapt to any natural language dataset and can work on any text-based deep learning models.

\section*{Impact Statement}

Understanding how neurons work is a step toward understanding how deep neural networks function. We believe that our work addresses some fundamental blockades in explainability research: the dependence on manual labor, and limited scalability.
Especially with the flooding number of LLMs and critical tasks that they might be applied on, understanding such models will be even more important.

We plan to open-source the code with the camera-ready version of this paper once it is accepted so they can be tested and understood better.
We hope that our work inspires more research in this direction.

\section*{Ethics Statement}

To our knowledge, this research has a very low risk of misuse potential, and it does not collect data from users while it relies on an existing text corpus and models that can be derived from people's writing activities.
This research does not include a human study.
The manual, blind annotation was performed by the authors.
All the outputs for ChatGPT were obtained manually by accessing their website by the authors. FlanT5-XXL model was run on NVIDIA Quadro RTX 8000 machines with 20 V-CPU cores and 40GB RAM. 

\section*{Acknowledgement}

This project was done as part of an industry-mentorship program at University of Massachusetts, Amherst. We would like to thank Prof. Andrew McCallum, TA Andrew Drozdov, and the PhD advisor for the project, Sheshera Mysore for their support and guidance throughout the course.


\bibliography{bibliography}


\clearpage
\appendix
\section{Descriptors Derived from AMZN} \label{appdx:desc}

\tab~\ref{tab:all_descriptors} lists all 26 descriptors extracted from AMZN dataset that activates neurons in the model as discussed in \sec~\ref{sec:method:candidates}.

\begin{table}[!ht]
\begin{center}\small
\captionof{table}{The 26 descriptors that we generated from AMZN dataset.} \label{tab:all_descriptors}
\begin{tabular}{|p{0.8in}| p{0.8in}|p{0.8in}|}
 \hline
 Age Appropriate / For Kids & Audio / Sound & Battery / Charging \\ 
 \hline
 Beverage & Cleaning / Maintenance & Color \\  
 \hline
 Controls & Design / Looks / Appearance & Durability \\
 \hline
 Fabric & Gift / Present & Graphics\\
 \hline
 Grip & Healthy / Fresh & Negative\\
 \hline
 Packaging\,/ Shipping\,/ Delivery & Positive & Price \\
 \hline
 Product Quality & Protection\,/ Safety & Size / Fit\\
 \hline
 Skincare / Haircare & Smell\,/\,Fragrance / Odor & Taste / Flavor\\
 \hline
 Texture & User Experience &\\
 \hline
\end{tabular}
\end{center}
\end{table}

\section{Correlation of Descriptors} \label{appdx:corr}

\begin{figure*}[b]
\centering
\includegraphics[width=\textwidth]{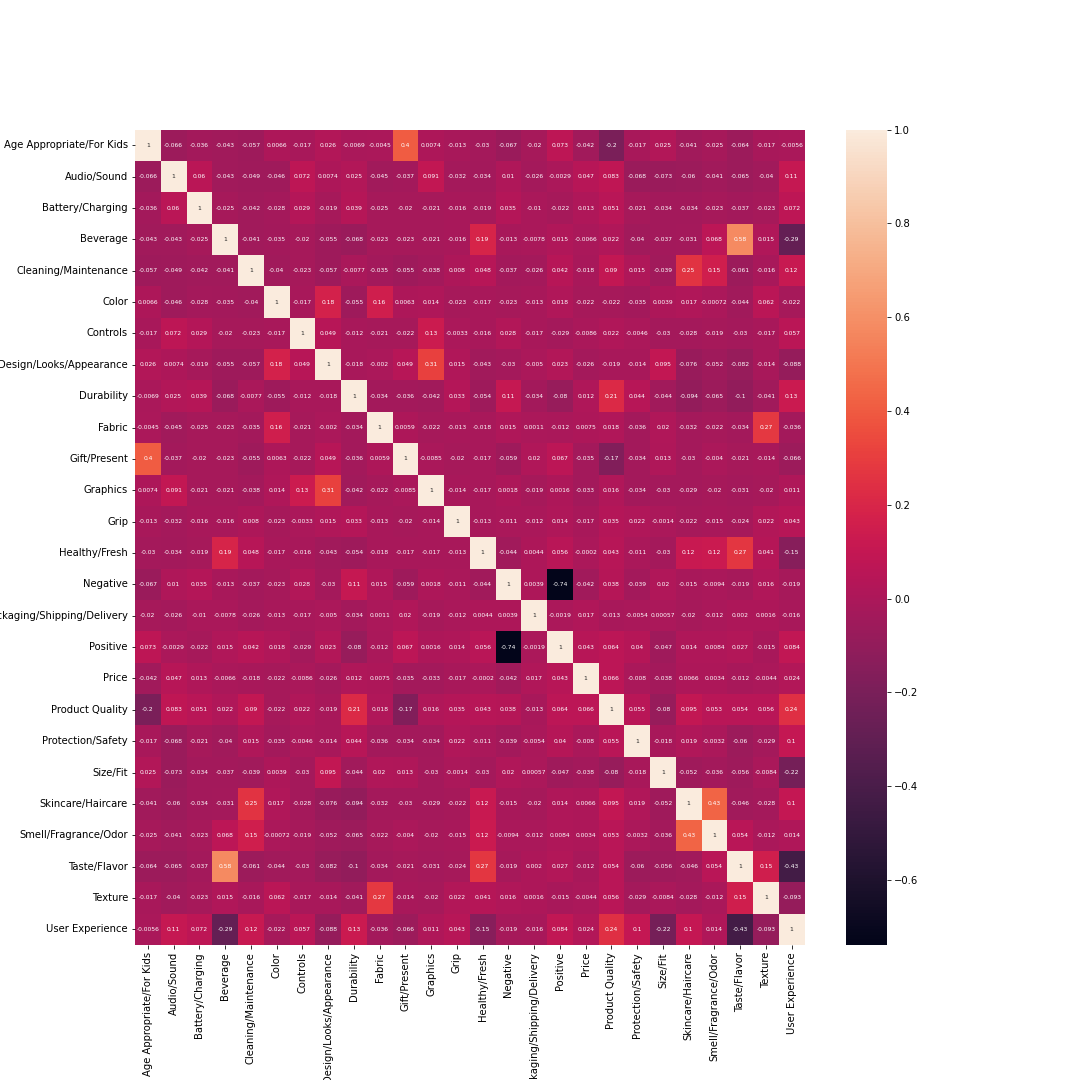}
\caption{Heat map of the correlation between 26 descriptors}
\label{fig:Concept_correlationMatrix}
\end{figure*}

\fig~\ref{fig:Concept_correlationMatrix} shows the correlation matrix of descriptors based on the response matrix from FLAN-T5 XXL on the sentences and the descriptors.

\section{Data and Descriptor Distribution} \label{appdx:dist}
\begin{figure*}[ht]
\centering
\includegraphics[width=\textwidth]{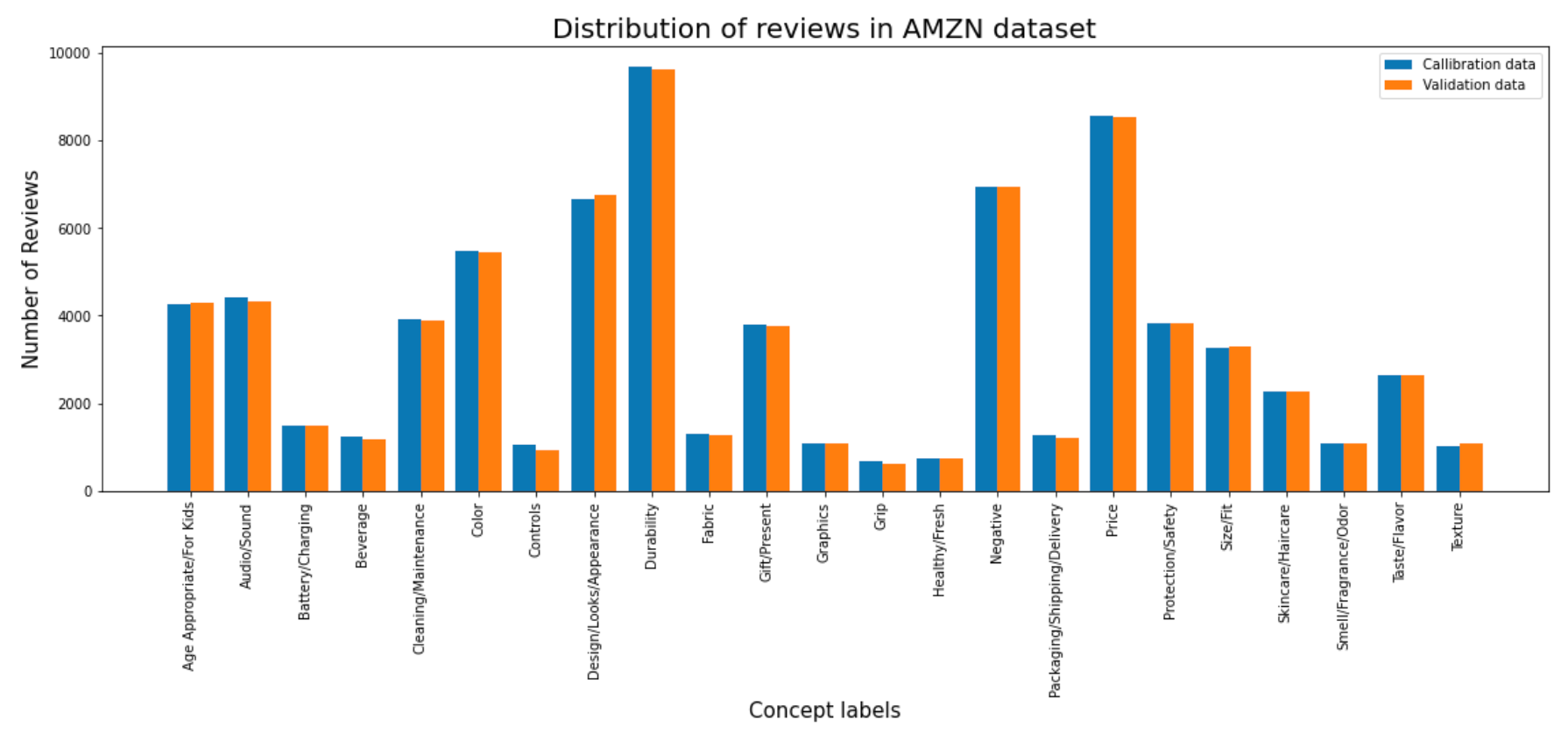}

\caption{Calibration and Validation subsets of AMZN dataset contain similar distribution across all the descriptors}
\label{fig:DataDistribution}
\end{figure*}

\fig~\ref{fig:DataDistribution} shows the distribution of data across different descriptors.

\begin{table*}[ht!]
\centering
\caption{Examples of descriptor generation using OpenAssistant SFT Pythia model and FlanT5 XXL model}
\small
\label{tab:concept-generation}
\begin{tblr}{
  width = \linewidth,
  colspec = {Q[180]Q[120]},
  vline{-} = {-}{0.08em},
  hline{-} = {-}{0.05em},
}
\textbf{Reviews} & \textbf{Descriptors generated from OpenAssistant and FlanT5 XXL model}\\
My son really likes the pink. Ones which I was nervous about & 	'user experience', 'color', 'pink', 'favorable' \\
I like this as a vent as well as something that will keep house warmer in winter.  I sanded it and then painted it the same color as the house.  Looks great. &  'looks great', 'as a vent', 'easy to use', 'user experience', hopeful' \\
Great book but the index is terrible. Had to write and high light my own cross ref info. & 'good', 'incomplete', 'well-written', 'user experience' \\
I recommend this starter Ukulele kit.  I has everything you
need to learn the Ukulele. &  'ample storage', 'easy to learn', 'comfortable', 'affordable', 'recommend', 'kit', 'ukulele' \\
The stained glass pages are pretty cool. And it is nice how the black outlines are super dark and thick. And that the dragons aren't all fighting with the wizards. & 'fun', 'user experience', 'gameplay', 'visuals', 'characters' \\
\end{tblr}
\end{table*}

\section{Examples of Descriptors and Clusters}\label{appdx:clusters}

\tab~\ref{tab:concept-generation} shows the 
examples of descriptors generated using Flan-T5 XXL and OpenAssistant SFT Pythia.
\tab~\ref{tab:examples_clustering} shows the 
examples of clustered descriptors. The representative descriptors are manually assigned for readability.

\begin{table*}
\centering
\caption{Examples of descriptors generated using Flan-T5 XXL and OpenAssistant SFT Pythia model clustered using agglomerative clustering and their final cleaned descriptors created manually.}
\label{tab:examples_clustering}
\small
\begin{tblr}{
  width = \linewidth,
  colspec = {Q[240]Q[80]},
  vline{-} = {-}{0.08em},
  hline{-} = {-}{0.05em},
}
\textbf{Generated descriptor} & \textbf{Final descriptor}\\
'yellow color', 'pretty colors', 'stunning color', 'unsatisfactory color', 'nicely colored', 'neon pink', 'pigments', 'redness', 'appealing color palette', 'richer pigment' & Color \\
'stench', 'bad scent', 'unexplainable chemical smell', 'fragrance selection', 'fragrance-free version', 'fragrance usage', 'nice smelling', 'smells delicious', 'different scents', 'pleasant aroma','scent control' & Smell/Fragrance/Odor \\






'fun for most ages', 'fun and challenging', 'enjoyed', 'good playing time', 'unexpectedly enjoyable', 'enjoyable. hopeful: nice', & Entertainment \\

\end{tblr}
\end{table*}

\end{document}